\documentclass[letterpaper, 10 pt, conference]{ieeeconf}  % Comment this line out if you need a4paper

\IEEEoverridecommandlockouts                              % This command is only needed if 
\usepackage{amsmath} % assumes amsmath package installed
\usepackage{algorithm}
\usepackage{algpseudocode}
\usepackage{graphicx}
\usepackage{color}
\usepackage{booktabs}
\usepackage{array}
\usepackage{svg}
\usepackage{multirow}
\usepackage{subfig}
\usepackage{afterpage}
\usepackage[skip=5pt,font=footnotesize]{caption}
\usepackage{setspace}
\usepackage{caption} 
\usepackage{amssymb} 
\usepackage{gensymb}

\PassOptionsToPackage{table}{xcolor}

\title{\LARGE \bf
Adaptive Anomaly Recovery for Telemanipulation: A Diffusion Model Approach to Vision-Based Tracking
}

\author{Haoyang Wang$^{1}$, Haoran Guo$^{1}$, Zhengxiong Li$^{2}$, \textit{Member, IEEE},\\ and Lingfeng Tao$^{3}$, \textit{Member, IEEE}% 
\thanks{*This work is supported by NSF \#2426469 and \#2426470.}% <-this % stops a space
\thanks{$^{1}$H. Wang and H. Guo are with Oklahoma State University, 563 Engineering North, Stillwater, OK 74078, USA (e-mail: haoyang.wang; haoran.guo@okstate.edu).}%
\thanks{$^{2}$Z. Li is with the University of Colorado Denver, Department of Computer Science and Engineering, 1380 Lawrence St. Center, LW-834, Denver, CO 80217, USA (e-mail: zhengxiong.li@ucdenver.edu).}%
\thanks{$^{3}$L. Tao is with the Kennesaw State University, Department of Robotics and Mechatronics Engineering, 840 Polytechnic Lane, Marietta, GA 30060 (e-mail: ltao2@kennesaw.edu)}%
}

\begin{document}

\renewcommand{\thefootnote}{\fnsymbol{footnote}}
\setlength{\skip\footins}{10pt} 

\AddToHook{cmd/@maketitle/after}{
  \begin{minipage}{\textwidth}
\vspace{0.1cm}
\centering
\includegraphics[width=1\textwidth]{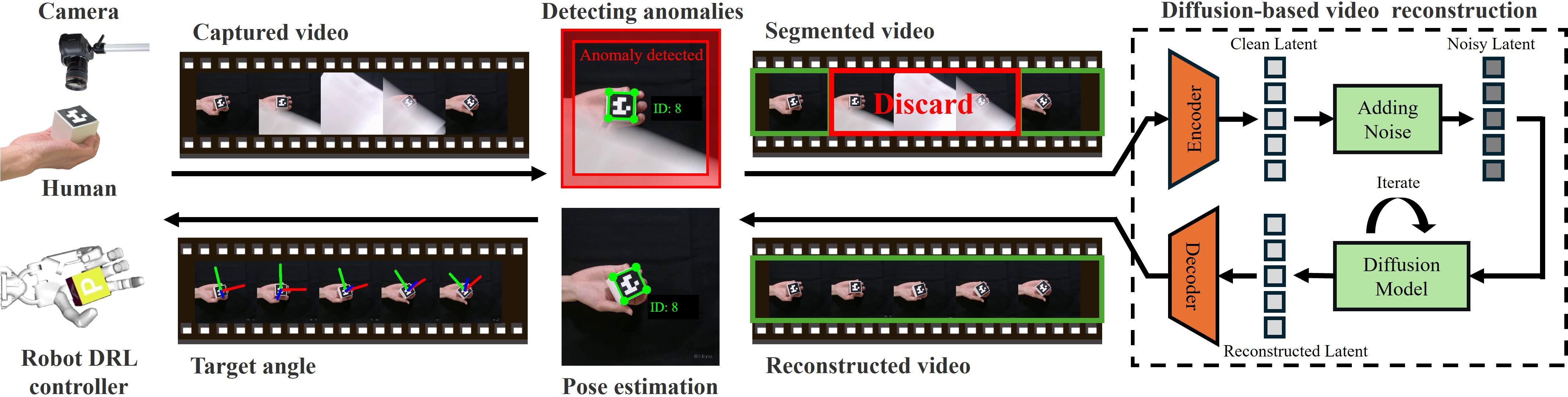}
\vspace{-0.4cm}
\captionof{figure}{
The system captures human motion via a camera, detecting and discarding anomalies using the FDD approach. The remaining frames undergo diffusion-based reconstruction, where an encoder maps them to a latent space, applies noise via forward diffusion, and restores them through reverse denoising. The reconstructed video is then used for pose estimation to determine target angles for the DRL-controlled robotic hand.
}
\vspace{-0.6cm}
\end{minipage}%
\par

}

\maketitle
\thispagestyle{empty}
\pagestyle{empty}

%%%%%%%%%%%%%%%%%%%%%%%%%%%%%%%%%%%%%%%%%%%%%%%%%%%%%%%%%%%%%%%%%%%%%%%%%%%%%%%%

%%%%%%%%%%%%%%%%%%%%%%%%%%%%%%%%%%%%%%%%%%%%%%%%%%%%%%%%%%%%%%%%%%%%%%%%%%%%%%%%

\noindent \begin{abstract}

Dexterous telemanipulation critically relies on the continuous and stable tracking of the human operator’s commands to ensure robust operation. 
Vison-based tracking methods are widely used but have low stability due to anomalies such as occlusions, inadequate lighting, and loss of sight.
Traditional filtering, regression, and interpolation methods are commonly used to compensate for explicit information such as angles and positions. These approaches are restricted to low-dimensional data and often result in information loss compared to the original high-dimensional image and video data.
Recent advances in diffusion-based approaches, which can operate on high-dimensional data, have achieved remarkable success in video reconstruction and generation.
However, these methods have not been fully explored in continuous control tasks in robotics.
This work introduces the Diffusion-Enhanced Telemanipulation (DET) framework, which incorporates the Frame-Difference Detection (FDD) technique to identify and segment anomalies in video streams. 
These anomalous clips are replaced after reconstruction using diffusion models, ensuring robust telemanipulation performance under challenging visual conditions. 
We validated this approach in various anomaly scenarios and compared it with the baseline methods. Experiments show that DET achieves an average RMSE reduction of 17.2\% compared to the cubic spline and 51.1\% compared to FFT-based interpolation for different occlusion durations.
\end{abstract}

%%%%%%%%%%%%%%%%%%%%%%%%%%%%%%%%%%%%%%%%%%%%%%%%%%%%%%%%%%%%%%%%%%%%%%%%%%%%%%%%
\vspace{-0.05cm}
\section{INTRODUCTION}
\vspace{-0.05cm}
%\noindent ~\cite{das2023review, su2023integrating}
\noindent Dexterous telemanipulation has emerged as a critical research area in robotics, especially for tasks that require fine remote manipulation, adaptive grasping, and dynamic adjustment to complex environments~\cite{darvish2023teleoperation}. Effective robot control depends on accurately translating human commands into robotic actions, typically through mapping human finger kinematics~\cite{meattini2022human} or tracking object motion~\cite{wang2024EFOLD}. These methods typically rely on vision-based tracking systems using RGB/RGBD and IR cameras (e.g., OptiTrack) to capture raw data, which is then processed by computer-vision techniques such as marker-based approaches (e.g., ArUco, AprilTag) or deep learning models like YOLOv5~\cite{kadam2024object}. However, they are vulnerable to occlusions, lighting variations, and human errors (e.g., accidental object drops), leading to information loss from the raw data that can disrupt control and compromise safety.

Researchers have explored various methods to address challenges associated with missing or unreliable visual information. A prevalent approach involves function approximation techniques, such as interpolation and regression, to predict motion trajectories based on historical data~\cite{luo2023clothoid}. Probabilistic filtering methods, including the Kalman filter, enhance estimation accuracy by integrating sequential measurements with the system~\cite{liu2022dynamic}. Additionally, learning-based approaches, such as multi-layer perceptrons (MLP) and long short-term memory (LSTM) networks~\cite{zhu2021learning}, leverage existing datasets to model operator preferences and predict future states. However, these methods primarily operate on post-processed, low-dimensional representations (e.g., position, velocity, and angular data), which inherently contain significantly less information than the raw data space, limiting their efficacy in high-dimensional perception and control tasks.

%High-dimensional data, such as images and videos, provide rich environmental context for complex dexterous manipulation tasks, enabling robotic systems to achieve a more comprehensive understanding of the real-time state space. 
High-dimensional data, such as images and videos, provide rich environmental context for complex dexterous manipulation tasks, enabling robots to understand the real-time state space better.
For instance, in a cube rotation task, raw visual data captures fine-grained finger gaiting movements and intricate object interactions, including changes in contact position and surface sliding. This detailed perception enhances the modeling of manipulation dynamics and supports more reliable control strategies. Consequently, when vision anomalies occur, it is crucial to repair or reconstruct the raw data space to preserve data fidelity. %\textit{That is to say, for dexterous telemanipulation tasks, it is essential to reconstruct the temporal and continuous video stream during anomalies to maintain operational stability and performance.} 
\textit{Thus, reconstructing the continuous video stream during anomalies is essential for stable telemanipulation.}
Traditional video reconstruction techniques, such as optical flow-based tracking and patch-based texture preservation methods ~\cite{li2017video, gu2019continuous}, have been widely used to restore missing visual information. However, these approaches often struggle with occlusions, rapid motion, and complex interactions between hands and objects, resulting in inconsistencies and loss of critical details essential for robust manipulation control.

Recent advancements in diffusion-based video generation models have shown notable improvements in frame inpainting, prediction, and reconstruction~\cite{chen2023seine}. These models iteratively refine noisy inputs into high-quality video sequences via large-scale datasets to learn spatiotemporal patterns, enabling smooth and coherent motion. Transformer architectures and attention mechanism further enhance their ability to capture long-range dependencies, ensuring structural consistency and motion continuity~\cite{liu2024sora}. While these models have proven effective in various visual processing tasks, their potential for predicting the trajectories of manipulated objects—and thereby enhancing the robustness of dexterous telemanipulation—remains largely unexplored.

In this work, we introduce the Diffusion-Enhanced Telemanipulation (DET) framework, which leverages a diffusion model for video reconstruction to enhance the robustness of telemanipulation. As illustrated in Fig. 1, a camera captures a video stream of the human operator's commands. To ensure the integrity of the video stream, anomalous events are identified and segmented using a Frame-Difference Detection (FDD) method. The remaining valid frames are then processed through a diffusion model, which reconstructs the missing sequences. The reconstructed video is used to infer object position and rotation, which serves as a real-time control target for a deep reinforcement learning (DRL) controller~\cite{wang2024EFOLD}, where telemanipulation is formulated as a Markov Game, modeling the human and the robot as agents interacting with each other. By integrating the DET framework, the DRL policy exhibits enhanced stability in achieving precise object positioning, significantly reducing the need for manual resets. While the proposed framework is demonstrated in a teleoperation task, it is readily adaptable to a broad range of vision-based tracking applications.

In particular, the contributions of this work are:

1). Propose a DET framework that merges FDD segmentation with diffusion-based video stream reconstruction, which enables high-dimensional motion prediction to enhance the robustness of telemanipulation under visual anomalies.

2). Validate DET within a Markov-Game-based telemanipulation environment to ensure stable, learning-based control.

3). Design comprehensive evaluation metrics to assess the performance of DET across various anomalies, command speeds, and durations while comparing baselines.

\section{RELATED WORK}

\subsection{Low-Dimensional Motion Prediction}

\noindent Current motion prediction approaches primarily rely on low-dimensional representations of system states, such as position, velocity, and orientation, to anticipate future trajectories. These techniques include interpolation and regression to fit observed data and predict future motion trends~\cite{liptaj2024general}. For instance, Luo et al.~\cite{luo2023clothoid} utilize a clothoid-based interpolation method for autonomous vehicle trajectory prediction, achieving significant improvements in handling sharp turns.
Probabilistic filtering methods, including the Kalman filter, enhance estimation accuracy by integrating sequential measurements with the system states~\cite{greenberg2020dynamic}. Liu et al.~\cite{liu2022dynamic} integrate Kalman filtering with IMU data, enabling robust dynamic state estimation in cluttered indoor environments. Deep learning methods have also received considerable attention. For example, Zhu et al.~\cite{zhu2021learning} introduce a recurrent neural network-based model to predict the motion of neighboring robots without relying on communication, thereby enhancing collision avoidance capabilities. However, these approaches typically focus on low-dimensional representations derived from raw sensor data, which omit rich visual details. Consequently, in high-dimensional perception and control scenarios, their performance is constrained.

\subsection{Video Reconstruction and Generation Method}
\noindent Traditional video reconstruction methods primarily rely on pixel-level computations (e.g., optical flow estimation and motion compensation). For example, Li and Cosker’s approach combines optical flow with Laplacian smoothness to enhance local flow estimation~\cite{li2017video}. Similarly, Gu et al. employ continuous bidirectional optical flow for video interpolation, leveraging bidirectional consistency to improve motion accuracy~\cite{gu2019continuous}. Beyond flow-based methods, learning-based frameworks such as Xu et al.~\cite{suzuki2020residual} use convolutional LSTM architectures for video restoration. Although effective, these methods often struggle to maintain stability and temporal coherence in complex, dynamic scenes, limiting their utility in real-time telemanipulation.

Recent advancements in diffusion-based models have significantly enhanced video reconstruction and generation by addressing temporal inconsistencies in dynamic scenes. For instance, diffusion techniques enable smoother frame interpolation by modeling complex motion patterns, leading to improved visual continuity~\cite{jain2024video}. Masked conditional frameworks further refine these predictions by effectively leveraging temporal context for both generation and interpolation tasks~\cite{voleti2022mcvd}. Moreover, the incorporation of transformer architectures into diffusion models, as exemplified by VDT, has enabled more robust spatiotemporal feature extraction and synthesis~\cite{lu2023vdt}. Complementing these technical strides, large-scale approaches have demonstrated the potential of video generation models as world simulators~\cite{Brooks2024VideoSim}, while open-source initiatives like Open-Sora democratize efficient video production for a broader research community~\cite{zheng2024open}.

\section{Diffusion-Enhanced Telemanipulation}
\vspace{-3pt}
\subsection{Anomaly Detection and Video Segmentation}
\vspace{-3pt}
\label{subsec:sora_framework}
\noindent In preliminary experiments, we observed that using diffusion models with the original video and text prompts often led to unstable anomaly corrections.
%In preliminary experiments, we observed that attempting to use the diffusion model to correct video anomalies with the original video alongside text-based prompts often resulted in unstable reconstructions. 
This instability is manifested as inconsistencies in object appearance, scale, and timing, particularly in longer video sequences or complex environments. 
In contrast, providing only the continuous video—such as frames preceding and following the anomaly—enabled more accurate reconstructions. 
%In contrast, providing only the continuous video—such as frames preceding and following the anomaly—enabled the model to generate more accurate missing frames. 
%In contrast, providing only the surrounding frames—before and after the anomaly—led to more accurate reconstructions.
Furthermore, the specific frame at which an anomaly is identified significantly influences the reconstruction outcome. 
For instance, if the input video contains a partial obstruction in motion, the model may inadvertently generate an extended occlusion process rather than restoring the intended scene.

To address these challenges, we developed the FDD method, which autonomously detects abrupt changes between normal and anomalous frames. By tracking variations at frame edges and analyzing the visibility of the region of interest (ROI)—in this case, the human hand and the object—the system effectively segments and filters out frames affected by occlusions, lighting variations, and other distortions. This ensures that only valid frames are retained for subsequent reconstruction, enhancing the robustness and reliability of vision-based telemanipulation.

In the FDD approach, the video stream is continuously monitored for anomalies (Algorithm~\ref{alg:intrusion_detection}).
We employ a dual detection strategy that combines frame-difference analysis and ROI tracking. ArUco marker detection enables ROI tracking in this work, though alternative tracking methods are also applicable.
Initially, the system captures the first frame as a baseline and defines an edge-region mask $M$ around the frame borders to detect intrusions (lines~1--3). ArUco marker detection parameters are set (line~4). Each incoming frame undergoes grayscale conversion, Gaussian filtering, and comparison with the baseline (lines~5--9). If pixel differences within $M$ exceed the threshold $P$, the frame is flagged as an intrusion candidate. Concurrently, ArUco markers are tracked (lines~10). If they vanish for $k$ or more consecutive frames, an anomaly is confirmed (line~11--12).
Upon detection, the anomalous interval $(S,E)$ is identified (line 15). The video is segmented into $Clip_A$ and $Clip_B$ to preserve valid frames while discarding corrupted ones (line 16). The processed clips are then sent to the Diffusion model for reconstruction.
The parameter values used in our implementation are $B = 50$, $T = 100$, $P = 1000$, and $k = 3$.

\newcommand{\Input}{\textbf{Input: }}
\newcommand{\Output}{\textbf{Output: }}

\begin{algorithm}[htp]
\makeatletter
\renewcommand{\ALG@name}{\scriptsize Algorithm}
\makeatother
\scriptsize \caption{\scriptsize Intrusion Detection and Video Segmentation}
\label{alg:intrusion_detection}
\begin{algorithmic}[1]

\Statex \hspace{-13pt} \Input Video sequence $V$, border width $B$, threshold $T$, pixel count threshold $P$
\Statex \hspace{-13pt} \Output Intrusion interval $(S, E)$, segmented video clips
\vspace{1pt}
\State Initialize video capture and read the first frame as the baseline
\State Convert baseline frame to grayscale and apply Gaussian blur
\State Define edge mask $M$ for detecting intrusions near the borders
\State Initialize ArUco marker detection parameters

\For{each frame $\mathbf{v}$ in $V$}
    \State Convert $\mathbf{v}$ to grayscale and apply Gaussian blur
    \State Compute frame difference with baseline
    \State Threshold the difference and apply mask $M$
    \State Count the number of abnormal pixels $D$ exceeds threshold $T$ within $M$ 
    \State Detect ROI in $\mathbf{v}$
    \If{no ROI is detected for $k$ consecutive frames \textbf{or} $D>P$} %$k$ consecutive frames no ROI detected \textbf{or} $D>P$
        \State Mark $\mathbf{v}$ as an intrusion frame
    \EndIf
\EndFor

\If{closed abnormal video clips detected}
    \State Define abnormal interval $(S, E)$ from the first to the last detected spot
    \State Segment video before $S$ to $Clip_A$ and after $E$ to $Clip_B$
\EndIf

\State \Return $(S, E)$, $Clip_A$, $Clip_B$
\end{algorithmic}
\end{algorithm}
\setlength{\textfloatsep}{0.1cm}

\vspace{-6pt}
\subsection{Diffusion-based Video Reconstruction}
\vspace{-3pt}
\label{subsec:diffusion_reconstruction}

\noindent Diffusion-based models have recently shown remarkable success in generating high-fidelity and coherent video sequences~\cite{zheng2024open}. In our DET pipeline, we leverage a pre-trained diffusion model to repair anomalous video segments identified by the proposed FDD module. Specifically, let $\{\mathbf{v}_1, \mathbf{v}_2, \dots, \mathbf{v}_T\}$ be the frames of the captured video, where each $\mathbf{v}_t \in \mathbb{R}^{H \times W \times 3}$ represents an RGB image with height $H$, width $W$, and three color channels. After FDD flags a subset of frames, denoted $\{\mathbf{v}_S, \dots, \mathbf{v}_E\}$, as anomalous, we replace those frames with reconstructed frames $\{\hat{\mathbf{v}}_S, \dots, \hat{\mathbf{v}}_E\}$ produced by the diffusion model.

To exploit the generative capability of diffusion, each image $\mathbf{v}_t$ is first encoded into a latent representation $\mathbf{z}_t = E(\mathbf{v}_t)$ via an encoder $E(\cdot)$. Next, the forward diffusion process incrementally adds noise to $\mathbf{z}_t$ over $L$ steps:
\begin{equation}\label{eq:forward_diffusion}
    q(\mathbf{z}_t^l \mid \mathbf{z}_t^{l-1}) = \mathcal{N}\bigl(\sqrt{1 - \beta_l}\,\mathbf{z}_t^{l-1},\,\beta_l \mathbf{I}\bigr),
\end{equation}
where $\mathbf{z}_t^l$ is the noisy latent at step $l$, and $\beta_l \in (0,1)$ is a schedule controlling the noise variance. During inference, the diffusion model $p_\theta$ which have learnable parameters $\theta$ iteratively denoises the latent from $\mathbf{z}_t^L$ back to $\mathbf{z}_t^0$:
\begin{equation}\label{eq:reverse_diffusion}
    p_\theta(\mathbf{z}_t^{l-1} \mid \mathbf{z}_t^l) = \mathcal{N}\!\bigl(\boldsymbol{\mu}_\theta(\mathbf{z}_t^l, l),\,\boldsymbol{\Sigma}_\theta(\mathbf{z}_t^l, l)\bigr).
\end{equation}
After reconstructing the latent $\hat{\mathbf{z}}_t^0$, a decoder $D(\cdot)$ maps it back to the pixel space, yielding $\hat{\mathbf{v}}_t = D(\hat{\mathbf{z}}_t^0)$. 
By replacing all anomalous frames with these high-fidelity reconstructions $\{\hat{\mathbf{v}}_S, \dots, \hat{\mathbf{v}}_E\}$, the overall video stream exhibits continuous and smooth transitions even under challenging conditions.

\vspace{-5pt}
\subsection{Multi-agent Modeling of DRL-based Telemanipulation}
\vspace{-5pt}

\noindent DRL methods have demonstrated their capability to handle dexterous manipulation such as rotating a block to a goal pose~\cite{andrychowicz2020learning} or solving a Rubik’s cube~\cite{akkaya2019solving}. We start with the mathematical modeling of the DRL-based telemanipulation approach. Typically, a single-agent DRL problem is modeled as a Markov Decision Process (MDP), which is defined as a tuple $\{S, A, R, \gamma\}$, where $S$ is the state space of the environment, $A$ is the set of available actions, $R: S \times A \rightarrow \mathbb{R}$ is the reward that is returned by the environment, and $\gamma \in [0,1]$ is the discount factor. The purpose of DRL training is to maximize the reward during the task. Unlike single-agent tasks, telemanipulation tasks involve two agents: a human operator and a robot. Thus, we model the telemanipulation problem as a Markov Game, an extension of MDP with multiple agents. Specifically, a telemanipulation task contains two MDPs shown in Fig.~\ref{fig:rl}. Each agent’s behavior follows an MDP and has its policy: the human operator as an agent, who is interacting with the object with state transition denoted as $\{H_t \rightarrow H_{t+1}\}$; and an autonomous agent following the human’s command to manipulate the object. The Markov game is a tuple $\{H, S, A, R, \gamma\}$.

\begin{figure}[t]
    \centering
    \vspace{2pt} 
    \includegraphics[width=6.3cm]{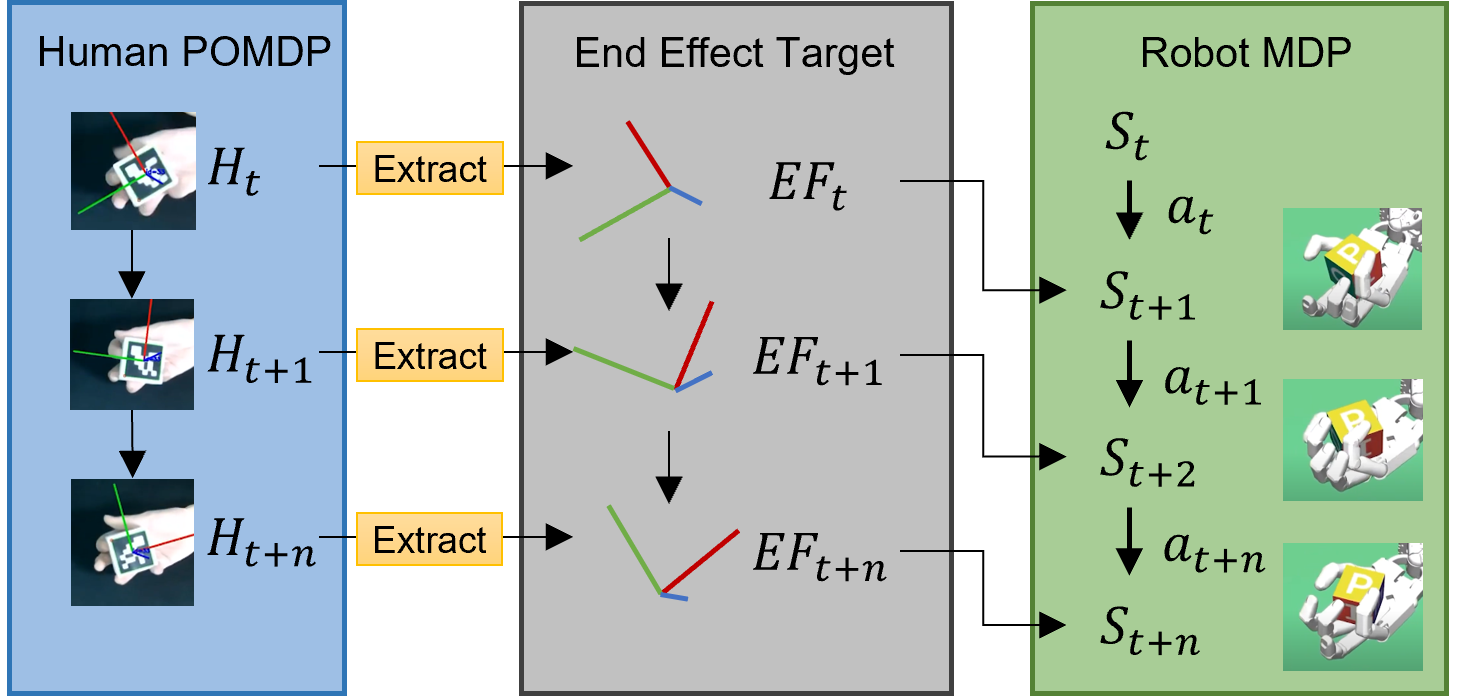}
    %\vspace{-5pt}
    \addtocounter{figure}{-1} % 先减少编号
    \caption{We model the telemanipulation tasks as a Markov Game that involves two agents: a human operator and a robot. The robot’s MDP depends on the operation command from the state of the human’s POMDP because the human’s mind is a black box and only the end effects that the human applied to the object can be observed by the robot.}
    %\vspace{-0.5cm} 
    \label{fig:rl}
\end{figure}

During telemanipulation, the robot is following the human, so its policy $\pi_{\mathrm{R}}: S \times H \times R$ depends on its state, the human command, and the reward. However, in the real world, the human mind is a black box, making the human MDP a Partially Observable MDP (POMDP)~\cite{spaan2012partially}, where the transition, action, and reward cannot be directly assessed, and only the human’s physical interaction with the object can be observed. We use discrete targets to train this policy, which ensures stable control even when the target changes. Combined with our proposed DET framework, the entire control process becomes more robust in complex environments.

\section{EXPERIMENTS AND EVALUATION METRICS}
\label{sec:experiments_and_metrics}

\subsection{Anomaly Type}

\noindent We identified four common types of anomalies—\textit{Occlusion}, \textit{Lighting Change}, \textit{Signal Noise}, and \textit{Out of Frame}—to evaluate their effects on our proposed DET framework.

\textbf{Occlusion} represents typical intrusions of objects into the camera view. We simulate this by adding a moving polygon into the video stream using OpenCV, ensuring that the timing and manner of each occlusion are controlled.

\textbf{Lighting Change} refers to cases where insufficient or excessive ambient light causes loss of crucial visual information. We implement this by decreasing the brightness of the video frame in OpenCV to simulate underexposure.

\textbf{Signal Noise} denotes general disturbance in the visual signal, which we simulate by gradually adding salt-and-pepper noise to the video frames in OpenCV.

\textbf{Out of Frame} refers to cases where the target object moves beyond the camera’s view. To ensure consistency, we recorded the video against a green screen and used OpenCV to shift the hand out of the frame programmatically.

\vspace{-5pt}
\subsection{Experiment Setup}

\begin{figure}[t]
    \centering
    \vspace{-1pt} 
    \includegraphics[width=7.8cm]{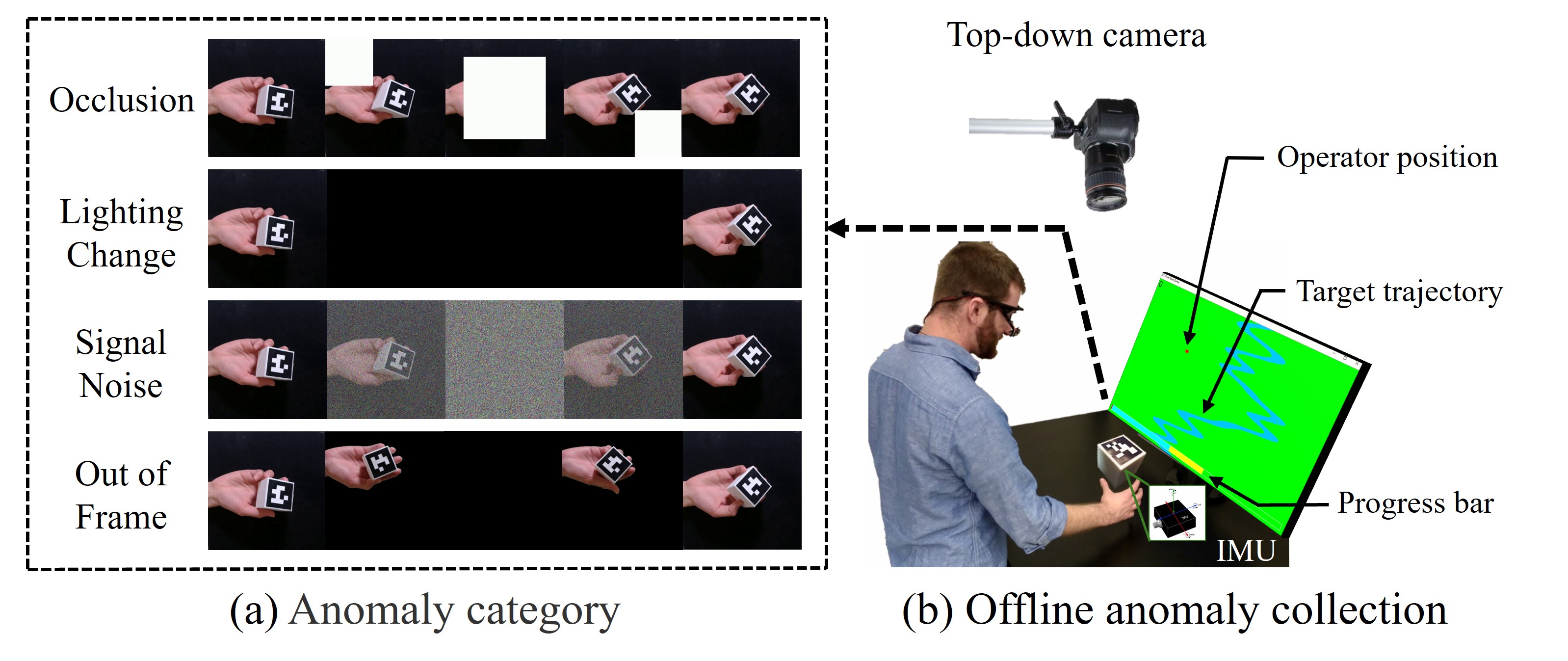}
    \vspace{-0.2cm}
    \caption{The offline data collection setup involves a human operator rotating a block with an ArUco marker, which is recorded by an overhead camera. The IMU provides ground truth rotation data. Added anomalies will disrupt the visibility of the marker.
    Then, the anomalous segments are reconstructed using the DET framework. These reconstructed frames are then used to control the robot in real time, and the resulting performance is compared against the ground truth data for evaluation.}
    %\vspace{-0.1cm}
    \label{fig:setup}
\end{figure}

\noindent We utilize \emph{Sora}—a state-of-the-art diffusion-based video generation model trained by large-scale data—as the reconstruction module in our DET framework~\cite{Brooks2024VideoSim}. 
However, since Sora is not currently open-source, data must be manually uploaded via its website for processing. To address this limitation, we developed a two-step validation process: 

\textbf{Offline data collection} (Fig.~\ref{fig:setup}). A human operator rotates a block following a sinusoidal or randomly generated trajectory, with the rotation angle displayed in real time (red dot) against a target trajectory (blue line). The sinusoidal trajectory is defined as $x_{t_\text{target}} = \mathcal{A} \sin(2\pi \gamma t)$, where $x_{t_\text{target}}$ is the target rotation angle, and $\mathcal{A} = 20^\circ$ denotes the maximum rotation to the left or right. For the random target, path points are randomly generated. A top-down camera records the block’s motion, while an onboard IMU logs its rotation and transmits the data for synchronized recording. Absent anomalies, the rotation angles from the camera and IMU agree and are denoted by $x_t$ at the $t$-th frame. When visual anomalies occur, the IMU measurements serve as ground truth. During offline data collection, we artificially introduce anomalies into the video via OpenCV and then reconstruct these segments using the DET framework. Each test video is constrained to 5\,s due to Sora's reference length limitation, recorded at 30\,FPS, in 720p resolution with a 720\,$\times$\,720 square format (i.e., $T = 150, H = 720, W = 720$). Anomalies of 0.5\,s, 1\,s, or 1.5\,s are inserted mid-video, with start/end frames $(S_A,E_A)$ set to $(68,83)$, $(60,90)$, or $(52,97)$, respectively.

\textbf{Real-time telemanipulation}. In this phase, the reconstructed videos are used to track the block's rotation angle in real time with the ArUco-based pose estimation. This angle is sent to the DRL-based controller as the manipulation target, which controls a Shadow Hand to track it in real time.

We designed three experiments: 
\textbf{Experiment 1} was designed to investigate different anomaly types, with the parameter set to $\gamma = 0.01$ and the anomaly duration fixed at 1 second. %记录五次操作员轨迹，每条轨迹对应一个重建后的视频，最后取平均
\textbf{Experiment 2} examined the effect of motion speed on anomaly perception. We defined four motion speeds (L1–L4), where L1 corresponds to slow, gentle rotations, and L4 involves rapid, large-amplitude movements. The corresponding parameters were set as $\gamma = 0.002$, $\gamma = 0.01$, and $\gamma = 0.03$ for L1–L3, respectively, while L4 followed a random trajectory. The anomaly duration remained 1 second for all conditions. %记录四次操作员轨迹，每条轨迹对应一个重建后的视频，最后取平均
\textbf{Experiment 3} explored the impact of anomaly duration under unpredictable motion conditions, using a random trajectory as the target. 
To collect statistical results, we recorded five different videos of the human operator for each anomaly type in experiments 1 and 3 and generated five different reconstructions for each of the videos with different speeds in experiment 2.

\subsection{Baseline Methods} %还可以压缩
\noindent In this section, we present two baseline methods from the low-dimensional domain processing for anomaly data reconstruction, providing a comparative foundation for evaluating different interpolation strategies.

\textbf{Fourier-based Method.} The Fourier-based approach transforms the known measurements into the frequency domain, reconstructs the signal via an inverse Fourier transform, and then fills any missing entries with the reconstructed values. To perform reconstruction, we apply the Discrete Fourier Transform (DFT) to the available sequence:
\begin{equation}
    F[k] = \mathrm{DFT}(x_t), \quad k = 0,1,\dots,N-1
\end{equation}
where \( F[k] \) represents the frequency components of the measured sequence \( x_t \), where $t\in[0, \dots, S_A-1]\cap[E_A, \dots, T]$. We then reconstruct the missing sequence by performing the inverse DFT (IDFT) on the filtered spectrum:
\begin{equation}
    \hat{x}_t = \mathrm{IDFT}(F[k]).
\end{equation}
where $\hat{x}_t$ represents the reconstructed element after filtering. Then used reconstructed sequence $\{\hat{\mathbf{x}}_{S_A}, \dots, \hat{\mathbf{x}}_{E_A}\}$ to fill in the anomaly part. This approach often excels at handling periodic or quasi-periodic signals, as the Fourier transform naturally captures dominant frequency components.

\textbf{Cubic Spline-based Method.} The goal of the cubic spline interpolation method is to estimate the missing values $\hat{x}_t$ for $t\in[S_A, \dots, E_A-1]$.
Cubic spline interpolation constructs a smooth piecewise cubic function that maintains continuity in both the function value and its first and second derivatives across segments. The interpolation process ensures that the reconstructed sequence smoothly connects with the available data points, avoiding abrupt changes.
To estimate $\hat{x}_t$, we solve a system of equations derived from these smoothness constraints. This system, typically represented in a tridiagonal form, determines the interpolation coefficients efficiently. Once solved, the missing sequence $\{\hat{\mathbf{x}}_{S_A}, \dots, \hat{\mathbf{x}}_{E_A}\}$ is reconstructed and replaces the anomaly-affected portion, providing a reliable estimate for non-periodic or irregular signals.

\subsection{Evaluation Metrics} 
\noindent To objectively evaluate the effectiveness of the reconstruction methods, we employ three quantitative metrics that assess both spatial and frequency-domain accuracy, as well as the smoothness of the reconstructed sequences.

\textbf{Root Mean Square Error (RMSE)}:
\begin{equation}
RMSE(x,\hat{x}) = \sqrt{\frac{1}{N}\sum_{i=1}^{N}(x_t - \hat{x}_t)^2}
\end{equation}
where $x_t$ represents the ground truth angle in degrees at the $t$-th frame, and $\hat{x}_t$ represents the measured angle of the corresponding reconstructed $t$-th frame $\hat{\mathbf{v}}_t$.
$RMSE$ reflects the magnitude of the overall deviation between two signals.

\textbf{Frequency Domain Shape Similarity Index (SSI)}:
\begin{equation}
SSI(f,\hat{f}) = \left( \frac{f}{\|f\|} \cdot \frac{\hat{f}}{\|\hat{f}\|} \right) \times 100
\end{equation}
where $f$ represents the result obtained by applying the Fourier transform to the ground truth data and taking the absolute value, and $\hat{f}$ represents the result obtained by applying the Fourier transform to the reconstructed sequence and taking the absolute value. In our implementation, we scale $SSI$ by 100 for interpretability.

\textbf{Endpoint Smoothness Index (ESI)}:
\begin{equation}
ESI = (1 - \frac{\alpha_S + \alpha_E}{2\pi}) \times 100
\end{equation}
where $\alpha_S$ and $\alpha_E$ represent the interior angles in radians formed between the beginning and end of the reconstructed segment and the adjacent non-anomalous data. Which quantifies the endpoint smoothness of each reconstructed sequence. For all measures, the lower is better.

\section{RESULTS AND DISCUSSION}
\subsection{Performance Under Different Types of Anomaly}
\renewcommand{\arraystretch}{1.2} % 适当缩小行间距
\setlength{\tabcolsep}{3.2pt} % 进一步压缩列间距

\begin{table}[b]
    \centering
    %\hspace{-10pt}
    \vspace{3pt}
    \caption{Performance Under Different Types of Anomaly}
    \vspace{-3pt}
    {\fontsize{6pt}{6pt}\selectfont % 指定表格字体大小
    \begin{tabular}{c c c c c c c c c c}
        \toprule
        \multirow{2}{*}{\textbf{Category}} 
        & \multicolumn{3}{c}{$\mathbf{Human}$} 
        & \multicolumn{3}{c}{$\mathbf{Robot}$}
        & \multicolumn{3}{c}{$\mathbf{GT-Robot}$}\\
        & \textit{RMSE(°)} & \textit{SSI} & \textit{ESI} 
        & \textit{RMSE(°)} & \textit{SSI} & \textit{ESI}
        & \textit{RMSE(°)} & \textit{SSI} & \textit{ESI} \\
        \midrule
        Occlusion       & 7.20  & 2.66  & 17.65  & 7.86  & 3.69  & 9.6 & 7.29  & 3.26  & 11.17 \\
        Lighting Change & 5.05  & 1.71  & 29.90  & 7.15  & 3.57  & 10.72 & 6.00  & 2.31  & 8.87 \\
        Signal Noise    & 5.83  & 2.00  & 38.96  & 7.37  & 3.63  & 6.89 & 6.25  & 2.76  & 8.10   \\
        Out of frame    & 7.50  & 2.58  & 16.94  & 10.47 & 2.41  & 12.35 & 6.74  & 2.56  & 11.50  \\
        \bottomrule
    \end{tabular}
    }
    \vspace{-17pt}
    \label{tab:scene_results}
\end{table}

In Table \ref{tab:scene_results}, we compare three entities: \textbf{Human}, representing the reconstructed human operator; \textbf{Robot}, the robot controlled by the reconstructed human operator; and \textbf{GT-Robot}, the robot controlled using ground-truth data. Across four anomaly scenarios, our proposed DET framework consistently demonstrates robust performance. In most cases, the Human entity achieves lower RMSE and SSI values compared to GT-Robot, indicating high reconstruction accuracy suitable for reliable robotic control. Furthermore, after applying DET, the Robot’s RMSE ranges between 7.15 and 10.47—closely aligning with GT-Robot—thus maintaining stable performance despite anomalous conditions. Although Human occasionally exhibits higher ESI, the Robot effectively compensates for these variations through its DRL policy, which is trained on discrete targets and provides substantial smoothing capabilities.

\subsection{Performance Under Different Object Motion Speed}

\renewcommand{\arraystretch}{1.2} % 调整行距
\setlength{\tabcolsep}{4pt} % 调整列间距
\begin{table}[b]
    \centering
    \vspace{10pt}
    \caption{Performance Under Different Object Motion Speed}
    \vspace{-3pt}
    {\fontsize{6pt}{6pt}\selectfont % 指定表格字体大小
    \begin{tabular}{c c c c c c c}
        \toprule
        \multirow{2}{*}{\textbf{Category}} 
        & \multicolumn{3}{c}{$\mathbf{Human}$} 
        & \multicolumn{3}{c}{$\mathbf{Robot}$} \\
        & \textit{RMSE(°)} & \textit{SSI} & \textit{ESI} 
        & \textit{RMSE(°)} & \textit{SSI} & \textit{ESI} \\
        \midrule
        L1 & 1.18  & 0.27  & 21.70  & 1.19  & 0.11  & 7.37  \\
        L2 & 8.05  & 1.94  & 14.06  & 9.47  & 2.33  & 4.51  \\
        L3 & 13.41 & 3.01  & 18.83  & 15.57 & 3.33  & 1.01  \\
        L4 & 16.03 & 7.36  & 26.47  & 15.69 & 7.80  & 5.78  \\
        \bottomrule
    \end{tabular}
    %=\vspace{-5pt}
    \label{tab:motion_results}
    }
\end{table}

Table \ref{tab:motion_results} illustrates that as object motion speed increases, both RMSE and SSI also rise, indicating greater reconstruction difficulty. At the lowest speed (L1), the metrics are nearly ideal, demonstrating close alignment with the ground truth. In contrast, at the highest speed (L4), RMSE reaches 16.03 and SSI climbs to 7.36, revealing the challenge of reconstructing rapid movements under anomalous conditions. Nonetheless, ESI remains moderate (15.06–26.47), preserving smooth rotational trajectories. The robot consistently maintains RMSE and ESI similar to those of the human, indicating the controller remains stable across different speeds. Notably, L3 exhibits relatively lower SSI, possibly suggesting that its intermediate speed aligns best with the control policy’s smoothing capabilities.

\captionsetup[subfigure]{labelformat=parens, position=top, font=tiny, aboveskip=-5pt, belowskip=-5pt}
 
\begin{figure}[t]
    \centering
    %\caption{Comparison of DET methods under different Lighting Change durations.}
    \subfloat{\includegraphics[width=0.16\textwidth]{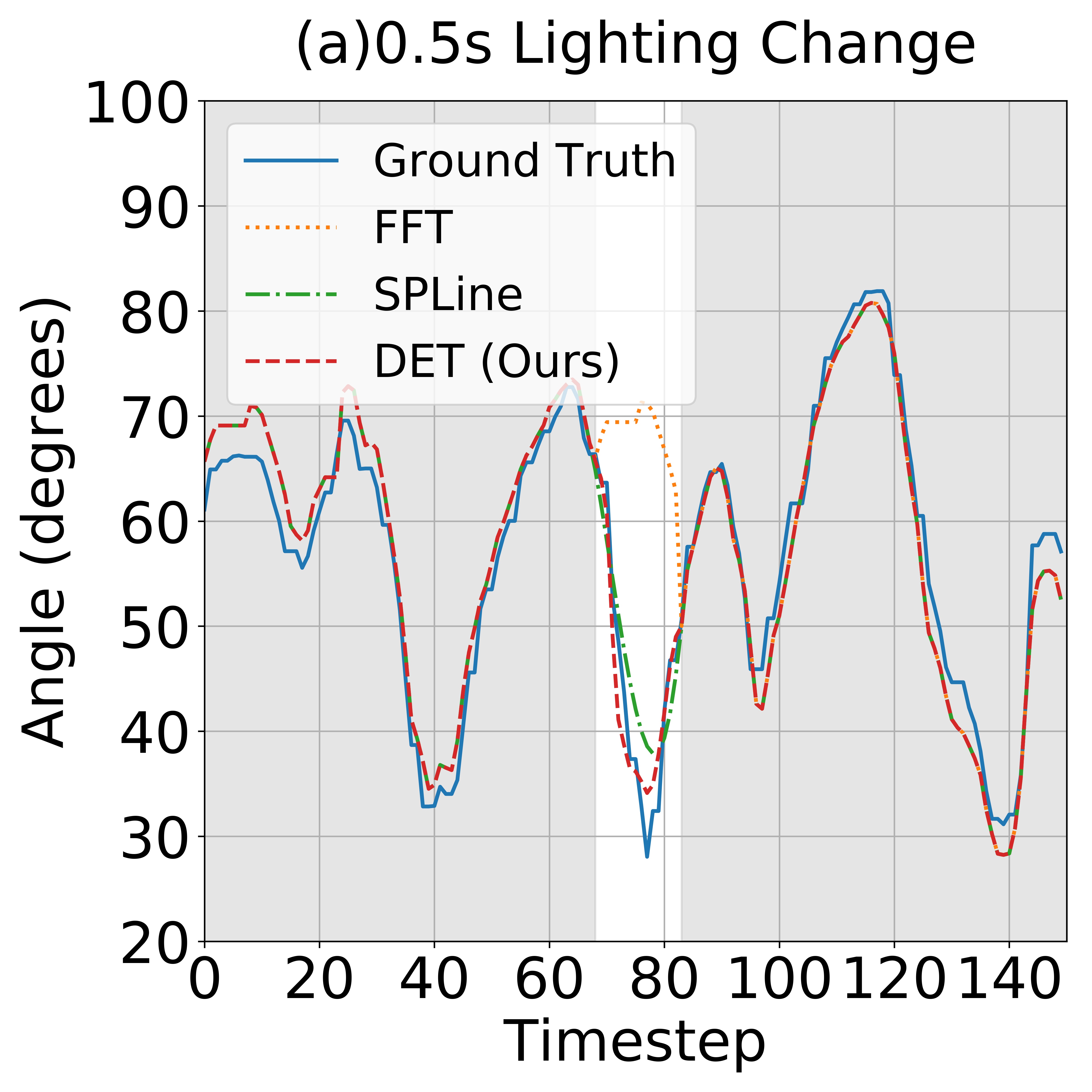}}
    \hfill
    \subfloat{\includegraphics[width=0.16\textwidth]{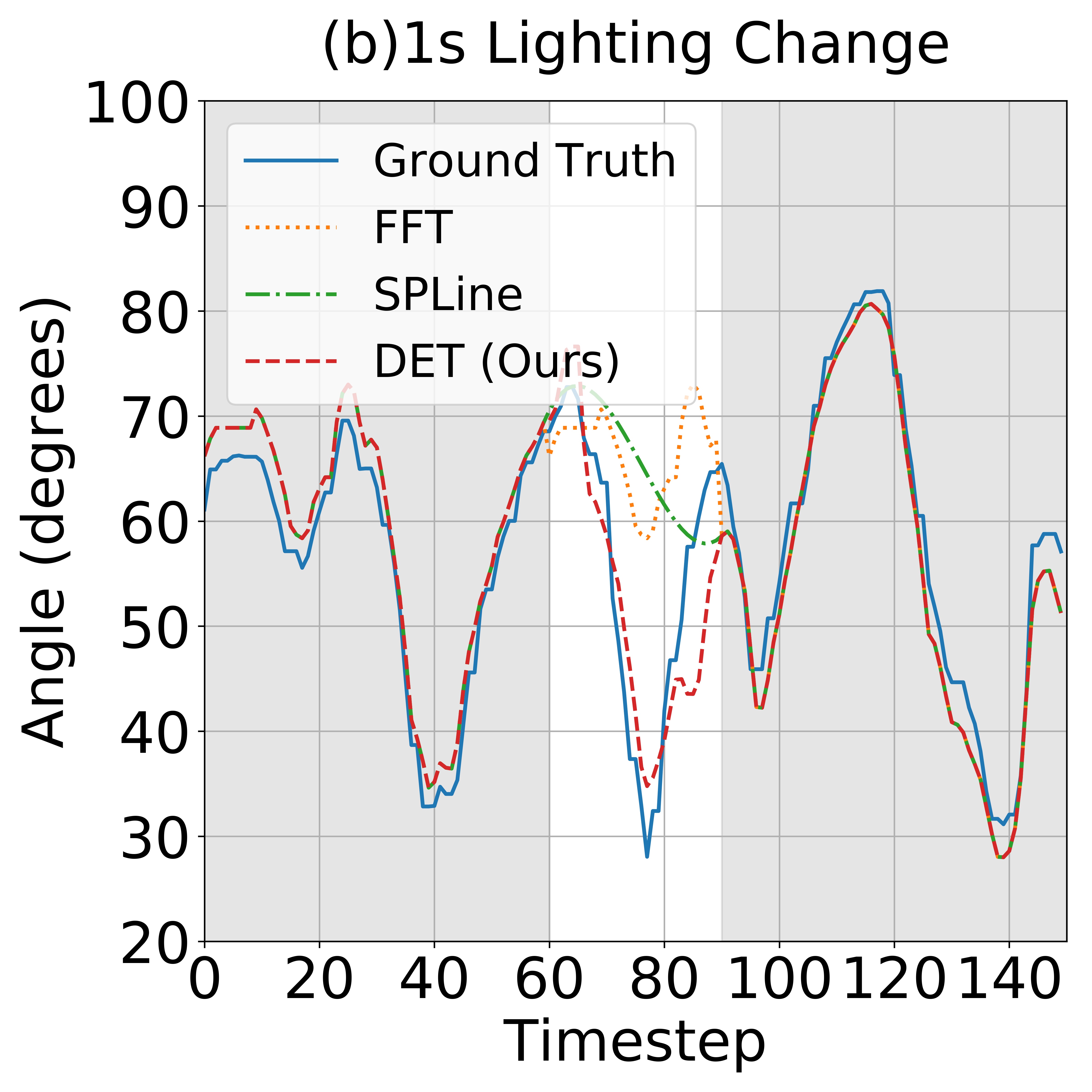}}
    \hfill
    \subfloat{\includegraphics[width=0.16\textwidth]{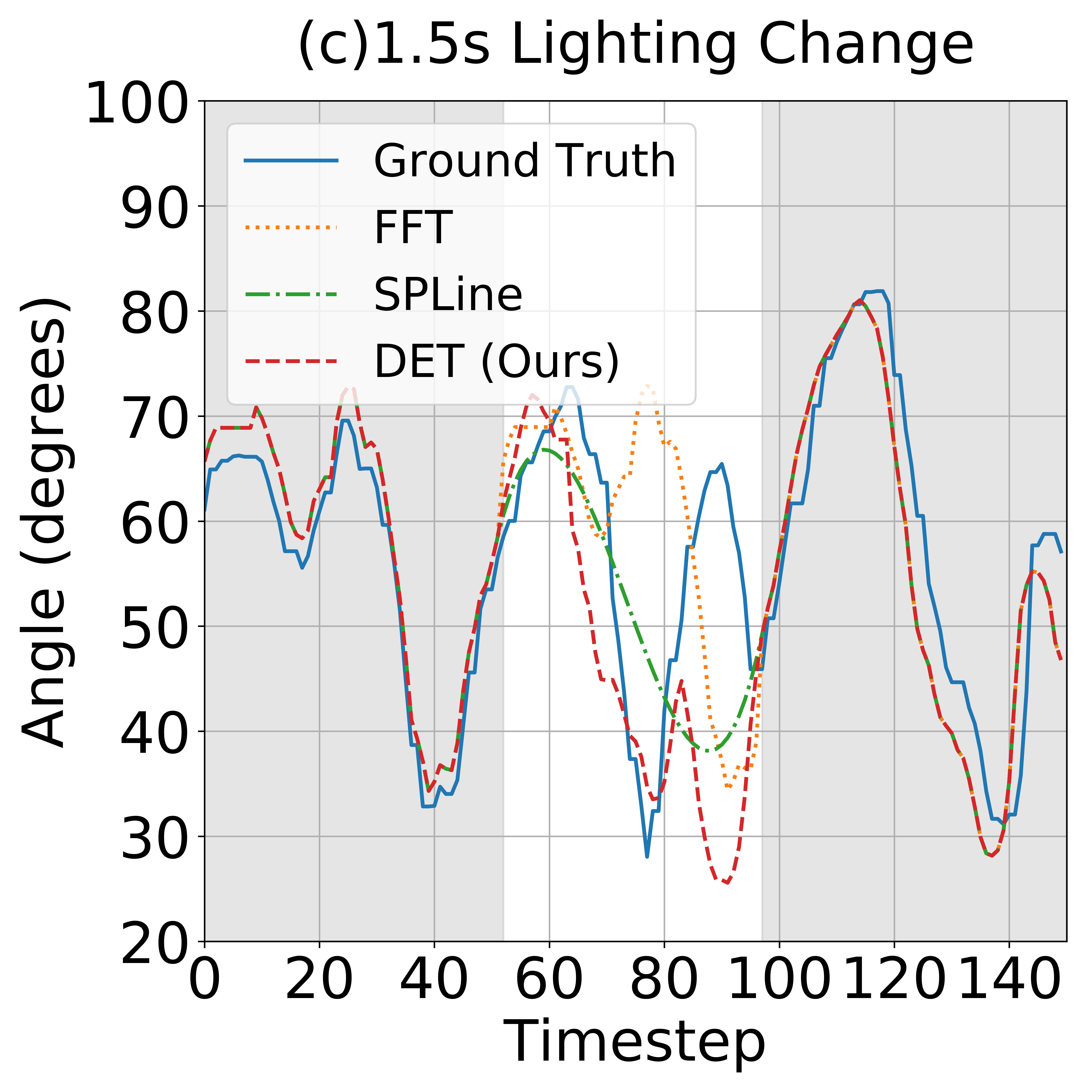}}
    \vspace{-4pt}
    \caption{Comparison of DET and baseline methods under different Lighting Change durations. DET achieved better performance in all conditions}
    \label{fig:three_images}
    \vspace{-5pt}
\end{figure}

\subsection{Performance Under Different Anomaly Durations and Compare with Baseline Methods}

\noindent Fig. \ref{fig:three_images} shows that DET (red) maintains better alignment with the ground truth throughout all Lighting
Changes (white area), whereas FFT (orange) introduces oscillatory artifacts due to its reliance on periodic signal reconstruction. The SPL (green) shows a smoother interpolation but struggles to maintain accurate alignment, particularly for longer occlusions.

\renewcommand{\arraystretch}{1.2} % 调整行距
\setlength{\tabcolsep}{2pt} % 调整列间距

\begin{table}[t]
    \centering
    \vspace{-4pt}
    \caption{Performance Under Different Anomaly Durations}
    \vspace{-2pt}
    {\fontsize{6pt}{6pt}\selectfont
    \begin{tabular}{c c c c c c c c c c}
        \toprule
        \multirow{2}{*}{\textbf{Category}} 
        & \multicolumn{3}{c}{$\mathbf{RMSE(\degree)}$} 
        & \multicolumn{3}{c}{$\mathbf{SSI}$} 
        & \multicolumn{3}{c}{$\mathbf{ESI}$} \\
        %\cmidrule(lr){2-4}\cmidrule(lr){5-7}\cmidrule(lr){8-10}
         & \textbf{\textit{Ours}} & \textit{FFT} & \textit{SPL} 
         & \textbf{\textit{Ours}} & \textit{FFT} & \textit{SPL} 
         & \textbf{\textit{Ours}} & \textit{FFT} & \textit{SPL} \\
        \midrule
        Lighting Change (0.5s) & 5.08 & 21.73 & 4.99 & 2.05 & 8.27 & 1.39 & 15.45 & 47.22 & 7.93 \\
        Lighting Change (1s)   & 11.01 & 24.64 & 15.97 & 5.45 & 6.40 & 8.04 & 10.30 & 51.71 & 5.36 \\
        Lighting Change (1.5s) & 17.27 & 21.80 & 19.31 & 6.57 & 3.36 & 6.97 & 10.87 & 42.60 & 6.78 \\
        \bottomrule
    \end{tabular}
    }
\end{table}

As shown in Table III, our method demonstrates competitive performance across varying Lighting Change durations, particularly achieving lower RMSE than FFT and SPL for longer anomalies (1s and 1.5s). Although SPL achieves marginally lower RMSE at the shortest duration, our approach consistently outperforms FFT in SSI and ESI, reflecting a robust capability to handle visual disruptions. Notably, while performance differences decrease as anomaly duration increases, our approach remains competitive, demonstrating robustness against longer anomalies. These results highlight the effectiveness of our method in handling varying anomaly durations in real-world scenarios.

In summary, the experimental findings confirm that the DET framework robustly handles diverse anomaly scenarios, including occlusion, lighting changes, signal noise, and out-of-frame motion, while the controlled robot maintains near-ground-truth performance.
The proposed approach consistently outperforms baseline methods (FFT and SPL) by achieving lower RMSE and smoother trajectories in most cases of varying anomaly durations.
Notably, its synergy with a DRL-based control policy effectively compensates for residual reconstruction smoothness, ensuring stable robotic operations despite significant target variations.
These results validate the DET framework’s feasibility in real-world conditions and highlight its potential for further optimization, particularly in high-dimensional data processing. Future work could focus on recognizing human hand skeletons and predicting human actions.
Although Sora does not currently provide a real-time API interface, the measured generation time after five tests for a 5-second 720P video is approximately 18.2 seconds, while for a 480P video, it is about 8.22 seconds. This implies that the diffusion-based model could enable real-time video reconstruction, especially if deployed locally in the future.

\bibliographystyle{IEEEtran} % 使用IEEEtran风格生成参考文献
\bibliography{ref.bib}

\end{document}